\documentclass[]{fairmeta}

\title{Skaling: Chinchilla's Exponents Meet Kaplan's Coupling}

\author[1]{Mathurin Videau}
\author[1]{Badr Youbi-Idrissi}
\author[1]{David Lopez-Paz}
\author[1]{Kartik Ahuja}

\affiliation[1]{FAIR at Meta}

\usepackage{amsfonts}
\usepackage[export]{adjustbox}
\usepackage[safe]{tipa} %
\usepackage{wrapfig}

\newcommand{\ska}{Skaling}

\abstract{
Neural scaling laws are foundational for language model development, yet standard formulations systematically under- and overestimate loss at data-scarce and overtraining extremes. This failure originates in the underlying assumption that model size and training data impact the loss independently. To address this, we introduce the \ska{} law (\textipa{/"skeIlIN/}), a generalized functional form that couples model capacity and data through a single interaction exponent. This simple extension reduces the Mean Absolute Percentage Error (MAPE) by $1.5$--$3\times$ across both interpolation and extrapolation regimes. When paired with a sparse grid strategy restricted to low-compute regimes, the \ska{} law achieves accurate full-grid extrapolation using approximately $10\times$ less compute than uniform sweeps. By enabling reliable performance prediction from small-scale experiments, the \ska{} law provides a more robust and resource-efficient framework for allocating compute budgets in next-generation model training.
}

\date{\today}
\correspondence{Mathurin Videau at \email{mvideau@meta.com}}

\begin{document}

\maketitle
\begin{figure}[!h]
    \centering
    \includegraphics[width=\linewidth]{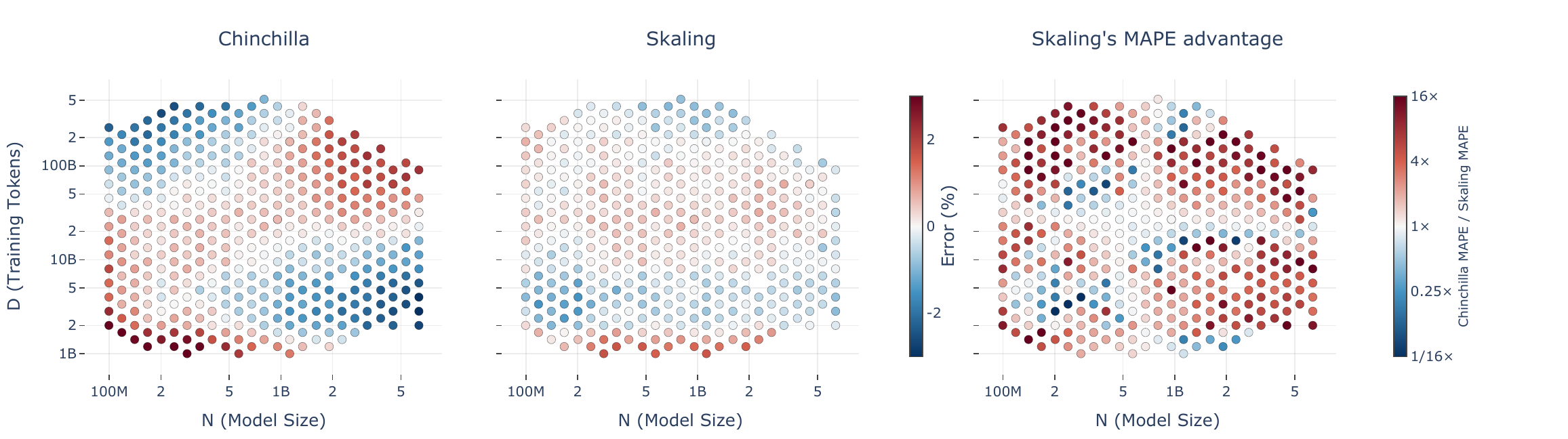}
    \caption{\textbf{The additive Chinchilla law carries a systematic, boundary-concentrated prediction bias that the \ska{} law removes.} Each marker is a trained $(N,D)$ configuration (model size $N$ horizontal, training tokens $D$ vertical). \emph{Left, centre:} signed percentage error (positive = overestimation (red), negative = underestimation (blue)) of the fitted Chinchilla and \ska{} laws (shared colorbar); Chinchilla shows a saddle-shaped residual that grows toward the corners, whereas \ska{} stays near zero throughout. \emph{Right:} the per-run ratio of the two laws' errors (capped at $16\times$); red runs where \ska{} is x times more accurate. \ska{} wins at $76\%$ of configurations (median $2.2\times$, and $\ge\!4\times$ at a third of them), with the largest gains at the cheaper edges.}\label{fig:model}
\end{figure}
\section{Introduction}\label{sec:intro}

The success of modern Large Language Models (LLMs) rests on their predictable and consistent improvement as training data and model size scale. This property is formalized by neural scaling laws \citep{kaplan2020scaling,hoffmann2022training,bi2024deepseek}, which predict the performance of high-compute runs using only a handful of low-compute models. Guided by these empirical laws, researchers can confidently optimize critical decisions regarding pretraining budgets, architectural dimensions, and resource allocation.

Despite their widespread adoption, standard scaling formulations possess a structural flaw~\citep{li2025predictable,ardalani2026how}. While the original Kaplan form~\citep{kaplan2020scaling} coupled model size and data, the widely used Chinchilla law~\citep{hoffmann2022training} models the reducible loss as a simple sum of independent terms for model size and training tokens. This mathematical structure implies a strict independence between the two variables, artificially enforcing a cross-derivative of exactly zero. However, early analysis \Cref{fig:model} shows Chinchilla is accurate in the interior of the grid but develops large, oppositely-signed errors toward the corners, reaching several percent where $N$ and $D$ are most imbalanced. This is the saddle-shaped residual expected when the $N$--$D$ interaction is omitted. Furthermore, our empirical analysis of the loss gradients reveals an interaction between these dimensions~(\Cref{sec:motivation}). Because they ignore this coupling, additive scaling models systematically under- and overestimate loss at the boundaries of the training grid.

To resolve this, we propose the \ska{} law, a minimal generalization of the Chinchilla form that introduces a single coupling exponent between model size and data volume, restoring the interaction of the Kaplan form that the additive Chinchilla law discards. Despite adding only one parameter, the \ska{} law reduces extrapolation MAPE by $1.5$--$3\times$ relative to Chinchilla across multiple cross-validation strategies. We further show that this improved functional form enables an efficient ``L-shape'' profiling strategy. By restricting training to the cheaper edges of the compute grid, sweeping data volume for small models (varying training tokens $D$, referred to as the D-band) and sweeping model size on a small, fixed amount of data (varying model size $N$, referred to as the N-band), the \ska{} law achieves predictive accuracy matching that of Chinchilla fitted on the full grid, while requiring up to $10\times$ less compute. Together, these contributions provide a practical, well-grounded framework for scaling decisions under constrained experimental budgets.

\textbf{Contributions}
\begin{enumerate}[label=\textbf{C\arabic*.}, leftmargin=2.4em]
    \item \ska{}: a coupled scaling form. We introduce a single interaction exponent between model size and data volume that corrects the boundary biases of additive laws (\Cref{sec:skaling_form}).
    \item Sparse profiling grids. We show that \ska{}'s predictive accuracy is preserved under a more effective sampling scheme, enabling ``L-shape'' grids that reduce profiling compute by $\sim\!10\times$ while maintaining accuracy (\Cref{sec:sampling}).
\end{enumerate}

\begin{figure}[!b]
    \centering
    \includegraphics[width=0.6\linewidth]{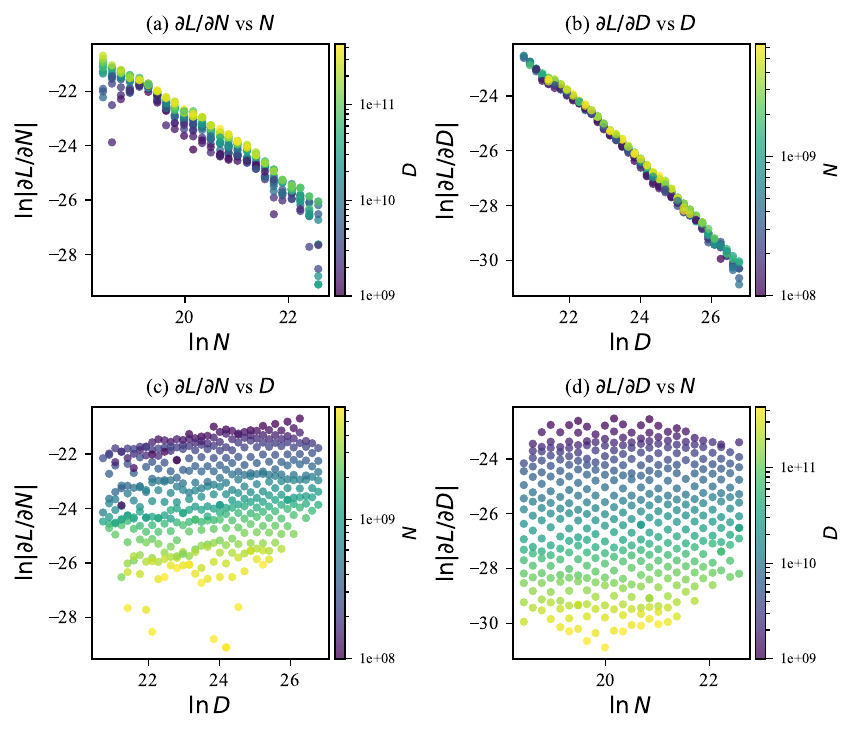}
    \caption{First-order derivative structure on Farseer (\Cref{eq:general_grad}; MLS estimates, log--log axes; the colorbar shows the cross-variable). \textbf{Top:} same-variable projections, $\ln|\partial L/\partial N|$ vs $\ln N$ (a) and $\ln|\partial L/\partial D|$ vs $\ln D$ (b), whose linear trends indicate power-law decay ($\alpha_N,\alpha_D$). \textbf{Bottom:} cross-variable projections, $\ln|\partial L/\partial N|$ vs $\ln D$ (c) and $\ln|\partial L/\partial D|$ vs $\ln N$ (d); the dominant structure is horizontal bands induced by the same-variable dependence, while the cross-slopes $\gamma_N,\gamma_D$ remain small.}
    \label{fig:deriv_deps}
\end{figure}

\section{The loss surface couples model size and data}
\label{sec:motivation}

Before committing to a functional form, we ask the data directly whether model size and training data interact. We probe the loss surface through its derivatives, estimated with the Moving least squares (MLS) \cite{lancaster1981surfaces} estimator (described in \Cref{app:numerical_gradients}); because the grid is logarithmically spaced, MLS returns log-slopes that we convert to real-space derivatives,
\[
\frac{\partial L}{\partial N}=\frac{L}{N}\frac{\partial\ln L}{\partial\ln N},\qquad
\frac{\partial L}{\partial D}=\frac{L}{D}\frac{\partial\ln L}{\partial\ln D},
\]
which removes the irreducible error $E$ and isolates the structure of the reducible loss.

We first summarize the first-order structure with the log-linear diagnostic
\begin{equation}\label{eq:general_grad}
\ln\!\left|\frac{\partial L}{\partial N}\right| = \alpha_N\,\ln N + \gamma_N\,\ln D + c_N, \qquad
\ln\!\left|\frac{\partial L}{\partial D}\right| = \gamma_D\,\ln N + \alpha_D\,\ln D + c_D,
\end{equation}
where $\alpha_N,\alpha_D$ capture the dominant same-variable decay and $\gamma_N,\gamma_D$ the residual dependence on the other axis; we use it only as a diagnostic, not as a scaling law. \Cref{fig:deriv_deps}(a,b) shows the same-variable projections are close to linear, with $\alpha_N\approx\alpha_D\approx-1.3$, so the marginal derivatives decay approximately as power laws. The cross-slopes are small, $\gamma_N\approx0.13$ and $\gamma_D\approx0.07$, so at first order the surface looks nearly separable; first-order projections alone, however, cannot rule out a weaker interaction.

\begin{figure}[!b]
    \centering
    \includegraphics[width=1.0\linewidth]{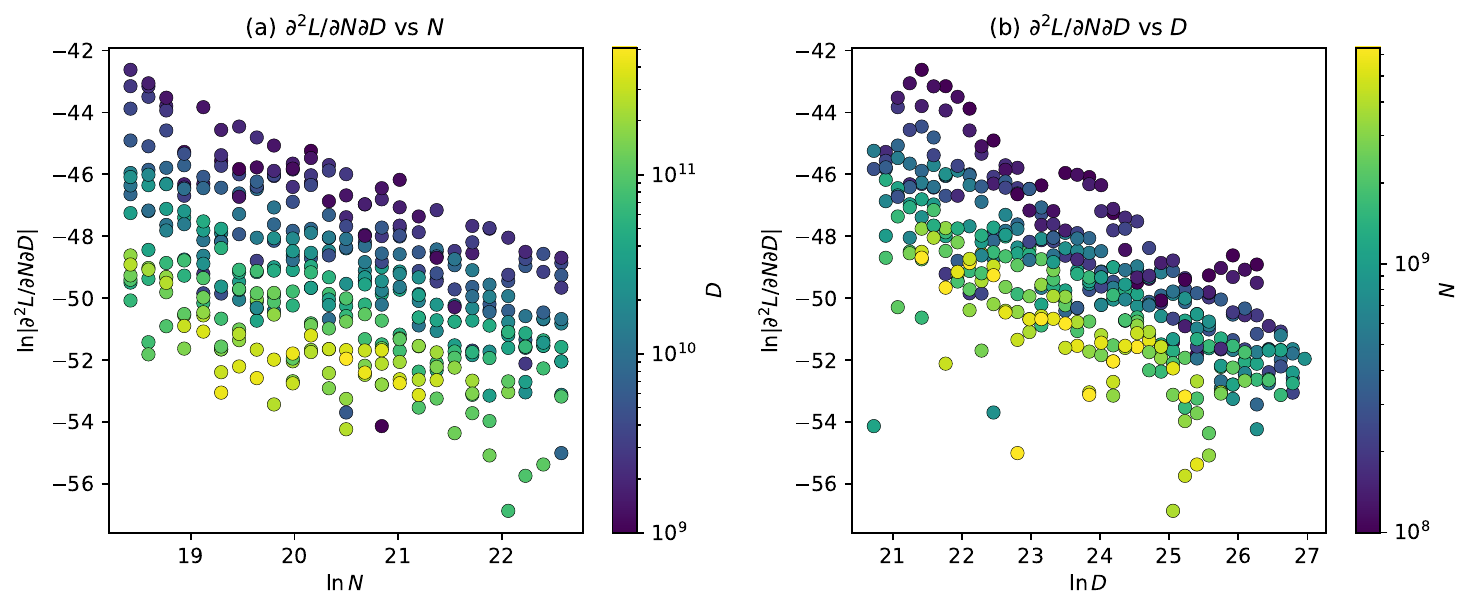}
    \caption{Cross-derivative $|\partial^{2}L/\partial N\partial D|$ (from the second-order term of the local quadratic fit, log scale), testing \Cref{eq:cross_deriv}. \textbf{(a)} Dependence on $\ln N$, colored by $D$. \textbf{(b)} Dependence on $\ln D$, colored by $N$. An additive law predicts $\partial^{2}L/\partial N\partial D=0$; the estimates are non-zero throughout the grid.}
    \label{fig:cross_deriv}
\end{figure}

The mixed derivative is the decisive test. Any additive law $L=f(N)+g(D)+E$ satisfies $\partial^{2}L/\partial N\partial D=0$ identically, whatever power laws are chosen for $f$ and $g$. \Cref{fig:cross_deriv} shows the estimated mixed derivative is instead non-zero across the entire grid, with its own power-law decay,
\begin{equation}\label{eq:cross_deriv}
\ln\!\left|\frac{\partial^{2}L}{\partial N\,\partial D}\right| = a\,\ln N + b\,\ln D + c,
\end{equation}
with $a\approx b\approx-1.1$ and a predominantly \emph{negative} sign: scaling $N$ and $D$ together lowers the loss more than scaling either alone, a synergy an additive law cannot represent. This motivates replacing Chinchilla's additive sum with a single coupling exponent, the \ska{} form introduced next; \Cref{app:cross_deriv} shows that this form reproduces both the negative mixed derivative and the small, asymmetric first-order slopes, and why a multiplicative coupling is preferable to an additive interaction term.

\section{Methods}\label{sec:method}
\subsection{The \ska{} form}
\label{sec:skaling_form}
Two functional forms anchor the scaling-law literature, differing primarily in how model size and data combine. The widely used Chinchilla law~\citep{hoffmann2022training} decouples them completely: it sums two terms with independent inner exponents $\alpha$ and $\beta$, alongside a free irreducible constant $E$. While mathematically convenient, this purely additive structure forces the cross-derivative $\partial^{2}L/\partial N \partial D$ to vanish identically, implicitly assuming that $N$ and $D$ do not interact. The earlier Kaplan form~\citep{kaplan2020scaling} takes the opposite approach: 
\[L(N,D)=\bigl[(N_c/N)^{\alpha_N/\alpha_D}+D_c/D\bigr]^{\alpha_D}\] 
Here the outer exponent $\alpha_D$ plays the role of \ska{}'s $k$, but Kaplan additionally ties the inner terms through the ratio $\alpha_N/\alpha_D$, so their per-axis decay rates are no longer independent.

\ska{} bridges these two paradigms. It retains Chinchilla's interpretable base terms and independent inner exponents but, following Kaplan, raises their sum to a single free outer exponent $k$:
\begin{equation}\label{eq:skaling}
    L(N,D) = \left(\frac{A}{N^{\alpha}} + \frac{B}{D^{\beta}}\right)^{k} + E
\end{equation}

This single parameter interpolates between the two prior forms. At $k=1$, the \ska{} law recovers the purely additive Chinchilla law. For any $k \neq 1$, it reinstates a Kaplan-style coupling and a non-zero cross-derivative. Crucially, unlike Kaplan, the \ska{} law achieves this coupling through the outer exponent $k$ while preserving Chinchilla's independent inner exponents. Despite this added flexibility, the \ska{} law retains two desirable properties of the additive law. First, because $k > 0$, the function remains \emph{strictly decreasing} in both $N$ and $D$: adding model capacity or training data never increases the predicted loss. Second, it preserves Chinchilla's structure. The exponent $k$ dictates how the two source terms \emph{aggregate} into the final loss, so each retains its individual interpretability even though their joint contribution no longer separates additively.

A practical payoff of this structure concerns compute allocation. Minimizing the loss under a fixed budget $C$ reduces to the same stationarity condition as the additive Chinchilla law, so the \ska{} law inherits Chinchilla's closed-form compute-optimal allocation: the optimal token-to-parameter ratio $D^{*}/N^{*}$ keeps its closed form and stays constant across scales when $\alpha \approx \beta$. Importantly, while the algebraic formula is identical, the resulting optimal ratio differs in practice because the underlying fitted parameters $A$, $B$, $\alpha$, and $\beta$ do not transfer between the two functional forms. We give the full derivation in \Cref{app:compute_optimal}. \Cref{app:compute_optimal} for the full derivation).

Accurately identifying this optimal ratio is critical because it dictates baseline architectural decisions. Frontier models such as DeepSeek~\cite{bi2024deepseek} typically lock in a fixed token-to-parameter ratio; therefore, the true optimum must be known to accurately quantify the performance penalty of suboptimal training. As detailed in \Cref{app:emp_opt_ratio} (\Cref{fig:opt_ratio}), the optimal ratio predicted by Chinchilla drifts significantly from that of \ska{}, accumulating to a 100-fold discrepancy at frontier compute scales. When validated against empirical measurements, \ska{}'s predicted ratio tracks the true underlying trend more accurately.

\subsection{Sampling Strategies}
\label{sec:sampling}
\begin{figure}[t]
    \centering
    \begin{subfigure}[b]{0.6\linewidth}
        \centering
        \includegraphics[width=\linewidth, trim={0 0 200 0}, clip]{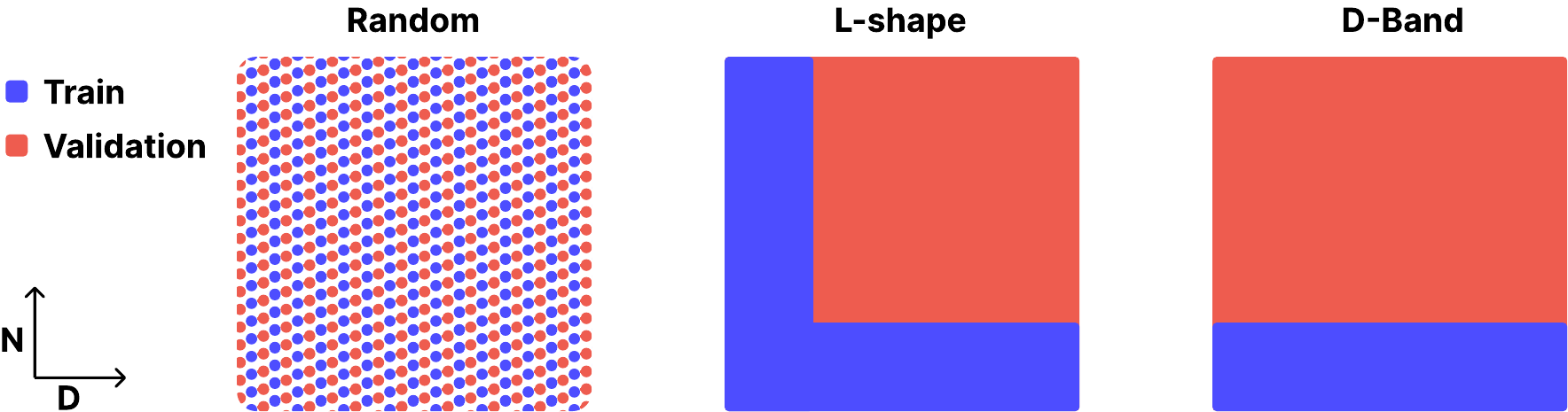}%
        \caption{Sampling strategies.}
        \label{fig:folds_band}
    \end{subfigure}
    \hfill
    \begin{subfigure}[b]{0.35\linewidth}
        \centering
        \includegraphics[width=\linewidth]{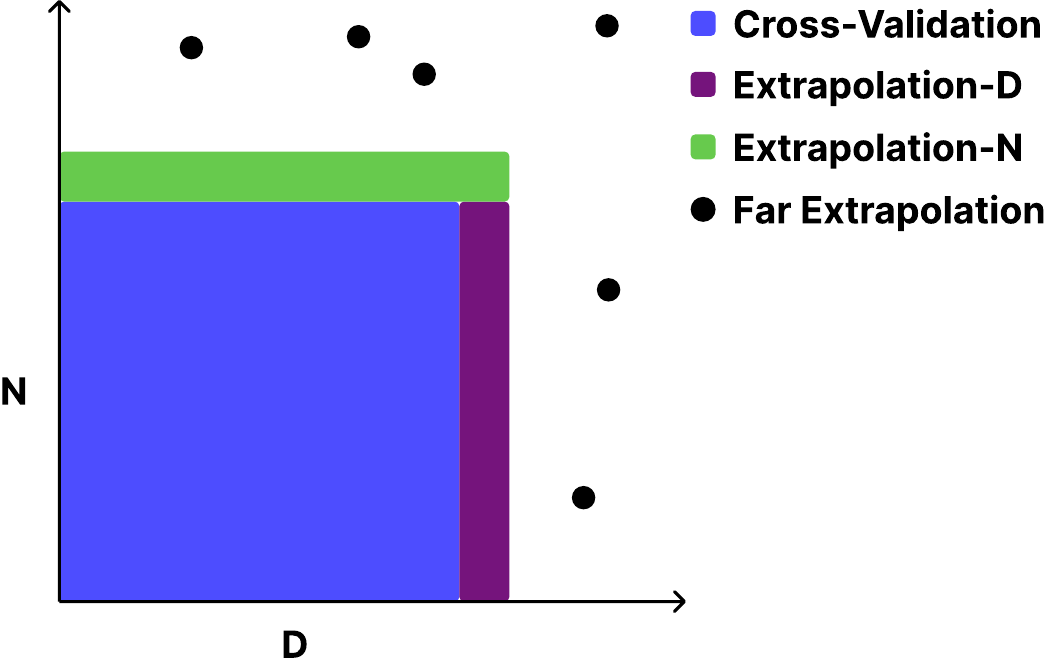}
        \caption{Evaluation regimes.}
        \label{fig:cv}
    \end{subfigure}
    \caption{Partitions of the $(N,D)$ grid. \textbf{(a)} Sampling strategies: \textbf{Random} spreads held-out points across the grid, whereas the \textbf{L-shape} grid restricts training to the low-compute edges. \textbf{(b)} Evaluation regimes used for cross-validation: interpolation, extrapolation in $N$ and in $D$, and far extrapolation beyond both.}
    \label{fig:splits}
\end{figure}

To obtain robust parameter estimates without prohibitive computational costs, the design of the experimental training grid plays an important role. Standard full-grid setups typically sample $N$ and $D$ on a square spaced logarithmically to capture behavior across multiple orders of magnitude. Because of this, the total compute is overwhelmingly dominated by the top-right corner of the grid, which contains the largest models trained for the longest horizons. This geometric concentration of compute makes dense sampling particularly wasteful. Rather than expending the vast majority of an experimental budget on a few massive runs, compute could be better allocated to a more discriminative set of points. The mathematical structure of the scaling laws naturally motivates our primary approach: the L-shape sampling strategy (\Cref{fig:folds_band}).

Consider the asymptotic behavior of the loss function. As data volume increases, the estimation error vanishes and the loss approaches the size-dependent approximation error:
\[ \lim_{D \to \infty} L(N,D) = \left(\frac{A}{N^{\alpha}}\right)^{k} + E \]
Consequently, increasing $D$ cleanly isolates the $N$-dependent coefficients. Symmetrically, as model size increases, the approximation error diminishes, cleanly isolating the data-dependent parameters:
\[ \lim_{N \to \infty} L(N,D) = \left(\frac{B}{D^{\beta}}\right)^{k} + E \]

In practice, we do not need to reach these theoretical limits. Simply varying one axis while holding the other fixed provides sufficient signal to trace the corresponding decay rate. The L-shape strategy applies this principle at the lowest compute scales. Rather than filling the entire parameter space, we sweep data volume $D$ exclusively for the smallest models to fit the data parameters ($B, \beta$), and sweep model size $N$ exclusively on the shortest training horizons to fit the size parameters ($A, \alpha$). By anchoring the independent decay rates along the grid boundaries, this sparse geometry maps the interaction between $N$ and $D$ more efficiently than a full grid sweep under same total compute budget constrain.

\subsection{Evaluation Protocol}\label{sec:evaluation_protocol}

To rigorously test the predictive capabilities and algorithmic stability of the fitted methods, we employ a comprehensive cross-validation framework. Rather than relying on a single static train-test split, we repeatedly resample the training data and construct corresponding hold-out sets. This approach allows us to explicitly quantify the uncertainty of our predictions, assess the variance of the fitted parameters, and test the models' ability to extrapolate reliably to unseen scales. For each cross-validation fold, the evaluation sets are partitioned as follows (\Cref{fig:cv}):
\begin{itemize}
    \item \textbf{Validation (Interpolation):} Randomly held-out points lying within the established boundaries of the training grid.
    \item \textbf{Extrapolation N:} Points at larger model sizes beyond the active training set, simulating prediction for larger architectures.
    \item \textbf{Extrapolation D:} Points at larger data volumes, testing predictions for extended training horizons.
    \item \textbf{Far Extrapolation:} The most challenging regime, consisting of the largest models trained on the largest data volumes, lying completely outside both the $N$ and $D$ boundaries of the training grid.
\end{itemize}

On all cross-validation folds, we report the MAPE on every evaluation set and the coefficient of determination ($R^2$) on the interpolation set only. For an evaluation set $\mathcal{S}$ in which run $i$ has measured loss $L_i$ and the fitted law predicts $\hat{L}_i$, the MAPE is the mean absolute relative deviation,
\begin{equation}
    \mathrm{MAPE}(\mathcal{S}) \;=\; \frac{100}{\lvert\mathcal{S}\rvert}\sum_{i\in\mathcal{S}}\frac{\bigl\lvert \hat{L}_i - L_i\bigr\rvert}{L_i}\quad[\%].
    \label{eq:mape}
\end{equation}
We restrict $R^2$ to interpolation because the extrapolation sets contain few points covering a narrow, somewhat arbitrary slice of the grid; since $R^2$ normalizes error by the variance of the held-out targets, it becomes unstable and uninformative there (often small or strongly negative), whereas MAPE stays directly comparable across all regimes. Aggregating over folds, we report the mean of each metric together with its variance.

\subsection{Compute extrapolation along iso-ratio slices}
\label{sec:isoratio}

\begin{wrapfigure}{r}{0.25\linewidth}
    \vspace{-\baselineskip}
    \centering
    \includegraphics[width=\linewidth]{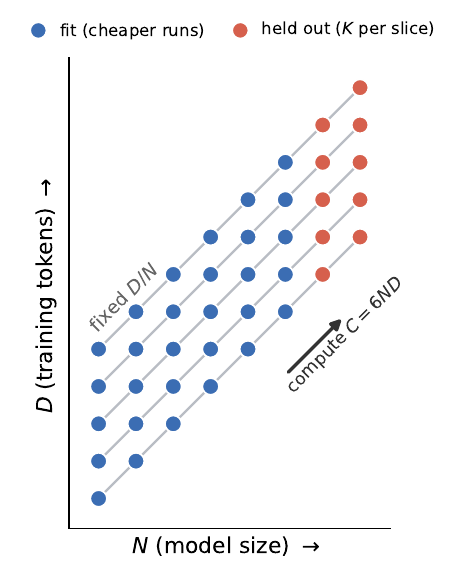}
    \caption{Iso-ratio compute extrapolation.}
    \label{fig:isoratio}
    \vspace{-\baselineskip}
\end{wrapfigure}

Frontier labs rely almost exclusively on one-dimensional compute power laws to project the performance of their massive pretraining runs. Because these models are typically scaled along a fixed token-to-parameter ratio, as demonstrated by DeepSeek~\citep{bi2024deepseek}, extrapolating compute along these specific operational rays represents an interesting point of comparison for any scaling formulation. While our earlier cross-validation comprehensively maps the $N$--$D$ grid, here we address a strictly operational challenge: can a single, globally fitted law reliably predict the most computationally expensive runs using only cheap, low-compute data?

To mirror this scenario, we group the runs into iso-ratio slices of constant token-to-parameter ratio $D/N$ (\Cref{fig:isoratio}). Within each slice, we hold out the $K$ highest-compute points ($K{=}8$). Every scaling law is refitted \emph{once} on the pooled remainder, the union of the low-compute points across all slices—and then evaluated on the held-out high-compute runs. This ensures each functional form sees the exact same training set for a rigorous global comparison. As a strong baseline, we additionally fit an independent single power law $L = A\,C^{a} + E$ strictly within each slice (\emph{per-ratio}), replicating the DeepSeek methodology~\citep{bi2024deepseek}. While this per-ratio baseline is heavily tailored to a single recipe and cannot inform joint $N$--$D$ allocation, it serves as an empirical upper bound on how well a dedicated, one-dimensional compute law can extrapolate along a fixed ray.

\section{Experiments}\label{sec:experiments}
\subsection{Experimental setup}

\paragraph{Data.} We fit and evaluate all scaling laws on two grids of pretraining runs.

Farseer~\citep{li2025predictable} records the final validation loss of $404$ $(N,D)$ configurations across $25$ model sizes ($100$M to $6.4$B parameters) and $55$ data budgets ($1$B to $512$B tokens), with compute ranging from $1.6\times10^{18}$ to $4.1\times10^{21}$ FLOPs. All models use a sequence length of $2048$. From this grid we hold out three evaluation sets: \emph{Extrapolation $N$} (the $3$ largest model sizes, $4.5$B--$6.4$B, $36$ points), \emph{Extrapolation $D$} (the top-$3$ data budgets per remaining model size, $66$ points), and \emph{Far extrapolation} ($7$ additional runs at larger scales, $2.3$B--$25$B parameters trained on $126$B--$453$B tokens, well beyond both axes of the training grid). The remaining $302$ configurations used for fitting total $\sim\!5.0\times10^{22}$ FLOPs.

SK-Grid, our own grid of training runs, is a complementary set of $134$ configurations across $15$ model sizes ($134$M to $4.9$B) and $16$ data budgets ($316$M to $316$B tokens), with compute from $9.0\times10^{16}$ to $9.9\times10^{20}$ FLOPs. The same hold-out scheme yields $7$ points for Extrapolation~$N$ ($2.8$B--$4.9$B), $33$ points for Extrapolation~$D$, and $3$ far-extrapolation runs at $\sim\!10^{22}$ FLOPs ($5.8$B--$10.8$B parameters). The fitting grid totals $\sim\!3.1\times10^{21}$ FLOPs.

\paragraph{Optimizer.} All scaling laws are fitted by minimizing a Huber loss in log space using L-BFGS-B with basin-hopping and autograd. Full optimizer settings and per-law bounds are given in \Cref{app:fitting}.

\paragraph{Baselines.} We benchmark the \ska{} law against the two most widely used reducible-loss forms: the additive Chinchilla law~\citep{hoffmann2022training} and the more heavily parameterized Farseer law~\citep{li2025predictable}. All laws are fit with the same optimizer and log-space objective (\Cref{app:fitting}), so that differences in predictive accuracy reflect the functional form rather than the fitting procedure. We fit each functional form to the remaining data and measure its predictive error under the cross-validation protocol of \Cref{sec:evaluation_protocol}.

\paragraph{Training.} The models in the grid are trained following the StepLaw hyperparameter prescriptions~\citep{li2025hyperopt}, which set the learning rate and batch size as functions of the model size and token budget. Using these near-optimal settings at every $(N,D)$ ensures that the measured loss reflects the scaling behaviour of the architecture rather than hyperparameter mistuning, providing a clean target for the scaling-law fits. Further details on model configurations and training hyperparameters are given in \Cref{app:pretraining}.

\subsection{Results}

\paragraph{Boundary errors.}
The clearest gains appear at the boundaries of the grid. Interior interpolation is already accurate for the additive Chinchilla law, but the error grows on the single-axis and far-extrapolation sets, where the saddle-shaped residual of \Cref{fig:model} is most pronounced. In \Cref{tab:cv_mape}, the \ska{} law reduces this boundary error consistently: on the full grids, the single-axis MAPE falls from $1.48$ to $0.47$ and from $1.98$ to $0.88$ on Farseer, and from $0.83$ to $0.39$ and from $1.44$ to $0.58$ on SK-Grid. The largest gains occur in the most imbalanced corners. On SK-Grid, for example, the far-extrapolation error drops from $5.17$ to $0.70$ on the full grid and from $14.63$ to $1.15$ on the L-shape grid; the largest-$N$ L-shape error drops from $6.09$ to $0.77$.

\paragraph{Sparse profiling.}
Sparse profiling preserves much more of \ska{}'s accuracy. When trained only on the L-shape grid, which uses roughly $10\times$ less fitting compute, the \ska{} law remains close to or better than the full-grid Chinchilla baseline on interpolation and single-axis extrapolation. Chinchilla, by contrast, degrades substantially under the same restriction: on Farseer, interpolation MAPE increases from $0.77$ to $2.51$, and on SK-Grid, far-extrapolation MAPE increases from $5.17$ to $14.63$. The coupled form therefore preserves predictive accuracy when the training grid is concentrated on the low-compute edges, which is the setting needed for the sparse profiling strategy in \Cref{fig:folds_band}.

\begin{table}[t]
\centering
\footnotesize
\setlength{\tabcolsep}{2pt}
\caption{Fit quality and predictive error on the Farseer and SK-Grid datasets. For each dataset we report the interpolation coefficient of determination $R^2$ ($\uparrow$) and the MAPE (\%, $\downarrow$; mean\,$\pm$\,std over 5 CV folds) on interpolation and three extrapolation regimes (larger model size $N$, larger data $D$, far extrapolation beyond both), for the full grid and the sparse L-shape grid (total training FLOPs per grid shown in the panel headers). $R^2$ is reported on the interpolation set only (see \Cref{sec:evaluation_protocol}). Best per column within each panel in \textbf{bold}.}
\label{tab:cv_mape}
\begin{tabular}{l ccccc @{\hspace{0.5em}} ccccc}
\toprule
 & \multicolumn{5}{c}{Farseer data} & \multicolumn{5}{c}{SK-Grid data} \\
\cmidrule(lr){2-6} \cmidrule(lr){7-11}
Law & $R^2$ & Interp. & Ext.\,$N$ & Ext.\,$D$ & Far & $R^2$ & Interp. & Ext.\,$N$ & Ext.\,$D$ & Far \\
\midrule
\textit{Full grid} & \multicolumn{5}{c}{$5.0\times10^{22}$~FLOPs} & \multicolumn{5}{c}{$3.1\times10^{21}$~FLOPs} \\
Chinchilla            & $0.995$ & $0.77{\pm}0.04$ & $1.48{\pm}0.03$ & $1.98{\pm}0.08$ & $2.46{\pm}0.19$ & $0.992$ & $0.81{\pm}0.14$ & $0.83{\pm}0.11$ & $1.44{\pm}0.03$ & $5.17{\pm}0.28$ \\
Farseer               & $0.982$ & $1.73{\pm}0.45$ & $2.37{\pm}0.08$ & $4.13{\pm}1.36$ & $2.43{\pm}1.93$ & $0.967$ & $1.66{\pm}0.32$ & $0.90{\pm}0.49$ & $4.45{\pm}0.21$ & $3.98{\pm}1.25$ \\
\ska{}                & $\mathbf{0.998}$ & $\mathbf{0.41{\pm}0.05}$ & $\mathbf{0.47{\pm}0.03}$ & $\mathbf{0.88{\pm}0.06}$ & $\mathbf{2.31{\pm}0.18}$ & $\mathbf{0.998}$ & $\mathbf{0.33{\pm}0.11}$ & $\mathbf{0.39{\pm}0.05}$ & $\mathbf{0.58{\pm}0.07}$ & $\mathbf{0.70{\pm}0.39}$ \\
\midrule
\textit{L-shape} & \multicolumn{5}{c}{$5.1\times10^{21}$~FLOPs} & \multicolumn{5}{c}{$6.5\times10^{20}$~FLOPs} \\
Chinchilla            & $0.954$ & $2.51{\pm}0.07$ & $4.32{\pm}0.13$ & $3.29{\pm}0.11$ & $9.82{\pm}0.48$ & $0.955$ & $2.19{\pm}0.10$ & $6.09{\pm}0.24$ & $3.63{\pm}0.13$ & $14.63{\pm}0.39$ \\
Farseer               & $0.974$ & $1.81{\pm}1.23$ & $2.07{\pm}1.72$ & $2.52{\pm}1.95$ & $2.37{\pm}1.33$ & $0.987$ & $0.82{\pm}0.52$ & $1.19{\pm}0.82$ & $2.66{\pm}2.42$ & $4.64{\pm}1.31$ \\
\ska{}                & $\mathbf{0.995}$ & $\mathbf{0.85{\pm}0.10}$ & $\mathbf{0.89{\pm}0.23}$ & $\mathbf{1.35{\pm}0.20}$ & $\mathbf{1.51{\pm}0.67}$ & $\mathbf{0.998}$ & $\mathbf{0.33{\pm}0.03}$ & $\mathbf{0.77{\pm}0.44}$ & $\mathbf{0.55{\pm}0.08}$ & $\mathbf{1.15{\pm}0.53}$ \\
\bottomrule
\end{tabular}
\end{table}

\paragraph{Interpolation is not sufficient.}
High interpolation fit quality is not enough to validate a scaling law. Chinchilla attains strong interpolation $R^2$ on the full grids ($0.995$ on Farseer and $0.992$ on SK-Grid), yet its extrapolation errors are several times larger than \ska{}'s at the grid boundaries. The relevant failure mode is therefore not a poor fit to the interior, but a systematic misprediction of how the loss surface bends away from the observed region.

\paragraph{Stable coupling.}
The fitted parameters explain why the boundary predictions improve. Across all grids in \Cref{tab:params_main}, the \ska{} law recovers a sub-unit coupling exponent, $k\approx0.31$--$0.45$, rather than collapsing back to the additive case $k=1$. The fitted irreducible loss is systematically lower than Chinchilla's, and on Farseer data it nearly vanishes: from $0.45$ to $0.03$ on the full grid and from $0.59$ to $0.05$ on the L-shape, while on SK-Grid it stays substantial ($1.75$ to $1.14$). We do not read $E\approx0$ as a vanishing loss floor. The two quantities that set how the loss flattens, the coupling $k$ and the floor $E$, trade off against each other: with $k<1$ the concave outer map makes the coupled reducible term decay more slowly at large scale, so it can absorb curvature that the additive law can only represent through a larger $E$. Because none of our runs reach the scale where the loss saturates, the data fix the total loss but not this split between a decaying term and a constant floor, and the \ska{} law resolves the ambiguity by pushing $E$ down, on Farseer data almost to zero. Beyond $E$, every parameter is determined precisely within each fit, with small fold-to-fold standard deviations: a few percent for the exponents and coupling, and at most $\sim\!40\%$ for the amplitudes.

\begin{table}[t]
\centering
\footnotesize
\setlength{\tabcolsep}{4pt}
\caption{Fitted coefficients (mean\,$\pm$\,std over $5$ CV folds) behind \Cref{tab:cv_mape}. Chinchilla is the $k{=}1$ special case of \ska{}; the nine-parameter Farseer law is omitted (see \citep{li2025predictable}).}
\label{tab:params_main}
\resizebox{\linewidth}{!}{%
\begin{tabular}{ll cccccc}
\toprule
Setup & Law & $A$ & $B$ & $\alpha$ & $\beta$ & $k$ & $E$ \\
\midrule
Farseer, full & Chinchilla & $(4.8{\pm}1.2){\times}10^{1}$ & $(1.1{\pm}0.1){\times}10^{2}$ & $0.27{\pm}0.01$ & $0.24{\pm}0.00$ & $1$ & $0.45{\pm}0.01$ \\
 & \ska{} & $(2.9{\pm}0.2){\times}10^{2}$ & $(6.0{\pm}0.3){\times}10^{3}$ & $0.32{\pm}0.01$ & $0.39{\pm}0.00$ & $0.41{\pm}0.01$ & $0.03{\pm}0.02$ \\
\addlinespace
Farseer, L-shape & Chinchilla & $(2.6{\pm}0.6){\times}10^{2}$ & $(1.0{\pm}0.1){\times}10^{2}$ & $0.39{\pm}0.02$ & $0.24{\pm}0.00$ & $1$ & $0.59{\pm}0.01$ \\
 & \ska{} & $(2.5{\pm}0.7){\times}10^{2}$ & $(1.7{\pm}0.4){\times}10^{3}$ & $0.32{\pm}0.01$ & $0.33{\pm}0.01$ & $0.45{\pm}0.03$ & $0.05{\pm}0.06$ \\
\addlinespace
SK-Grid, full & Chinchilla & $(5.0{\pm}1.2){\times}10^{2}$ & $(7.1{\pm}0.8){\times}10^{2}$ & $0.34{\pm}0.01$ & $0.31{\pm}0.01$ & $1$ & $1.75{\pm}0.02$ \\
 & \ska{} & $(5.3{\pm}1.8){\times}10^{6}$ & $(7.1{\pm}2.8){\times}10^{6}$ & $0.73{\pm}0.01$ & $0.63{\pm}0.01$ & $0.31{\pm}0.02$ & $1.14{\pm}0.06$ \\
\addlinespace
SK-Grid, L-shape & Chinchilla & $(1.0{\pm}0.0){\times}10^{4}$ & $(1.7{\pm}0.4){\times}10^{3}$ & $0.52{\pm}0.00$ & $0.36{\pm}0.01$ & $1$ & $2.16{\pm}0.02$ \\
 & \ska{} & $(1.0{\pm}0.0){\times}10^{7}$ & $(6.5{\pm}2.4){\times}10^{6}$ & $0.77{\pm}0.02$ & $0.63{\pm}0.01$ & $0.31{\pm}0.02$ & $1.18{\pm}0.12$ \\
\bottomrule
\end{tabular}}
\end{table}

\paragraph{Parameter count.}
The more heavily parameterized Farseer baseline does not remove the boundary failure by itself. Despite its additional degrees of freedom, it is less accurate than the \ska{} law in most regimes of \Cref{tab:cv_mape}, and its largest errors concentrate on data extrapolation (MAPE $4.13$ and $4.45$ on the Farseer and SK-Grid full grids). This suggests that the gain is not simply a consequence of adding parameters, but of using a functional form whose inductive bias matches the observed $N$--$D$ interaction. Part of the difference may also be that the richer form is harder to fit; we tried several optimizers, including Farseer's own pipeline, without obtaining a substantially better fit (\Cref{app:fitting}), so we report the best Farseer results we could obtain.

\paragraph{Coupling strength.}
The benefit weakens when the observed coupling is closer to additive. On Farseer-code and on the original Chinchilla measurements in \Cref{app:extra_datasets}, the fitted coupling is much closer to additive ($k\approx0.77$--$0.90$), and the \ska{} law correspondingly performs at roughly Chinchilla-level accuracy. This behaviour is expected from the nested form: because Chinchilla is recovered at $k=1$, the \ska{} law departs from the additive law when the data support a coupled surface and otherwise remains close to the additive fit.

\paragraph{Farseer data allocation frontier.}
On Farseer data, the same interaction has a concrete allocation consequence. Using numerical loss gradients recover token-to-parameter ratio that decreases with compute, with fitted exponents $-0.14$ and $-0.15$. This agrees with \ska{}'s analytic exponent ($-0.11$) and has the opposite sign from Chinchilla's near-flat prediction ($+0.03$). One order of magnitude beyond the data, the two prescriptions differ by roughly $10\times$ in the recommended token-to-parameter ratio (\Cref{fig:opt_ratio}).

\paragraph{Allocation direction.}
The sign of this allocation trend is dataset-specific rather than a universal consequence of the coupled form. For Farseer, the fitted \ska{} exponents satisfy $\alpha<\beta$, giving a decreasing analytic $D^\star/N^\star$ as compute grows. On SK-Grid, however, the fitted exponents satisfy $\alpha>\beta$ on both the full and L-shape grids, so the same closed-form optimum would increase the token-to-parameter ratio with compute. The robust conclusion is therefore that coupling changes large-scale allocation, while the direction of that change depends on the fitted data and architecture.

\subsection{Compute extrapolation}
\label{sec:results_isoratio}
The cross-validation above held out grid \emph{corners}; we now test the operational case of \Cref{sec:isoratio}: predicting the most computationally expensive runs of each recipe from computationally inexpensive ones. \Cref{tab:compute_extrap} reports the error on the $112$ held-out high-compute Farseer runs, split by training regime (the iso-ratio slices grouped into undertrained / optimal / overtrained thirds by $D/N$), with every law refit only on the computationally inexpensive remainder. \ska{} is the best \emph{global} law in every regime and overall (pooled MAPE $0.60{\pm}0.27\%$, a $3.9\times$ reduction over the additive Chinchilla law and below the far more heavily parameterized Farseer law), and the most stable---its error never exceeds $0.9\%$ in any regime. Chinchilla, by contrast, is strongly regime-dependent and weakest of all laws in the optimal band ($3.47\%$), so its pooled number hides where it fails. The only reference that edges the \ska{} law is the \emph{per-ratio} power law, and only near the optimum ($0.77$ vs $0.88\%$), where a one-dimensional law along a fixed ratio is naturally well behaved; it is fit separately per recipe and so cannot inform joint $N$--$D$ allocation. Chinchilla exceed $2.3\%$, confirming the gain comes from the functional form rather than the protocol.

\begin{table}[t]
\centering
\setlength{\tabcolsep}{4pt}
\caption{Compute extrapolation on Farseer (\Cref{sec:isoratio}) by training regime: iso-ratio slices grouped into undertrained ($D/N{=}1.8$--$7$), optimal ($10$--$40$) and overtrained ($56$--$158$) thirds, plus all $14$ setups. Predictive error on the $112$ highest-compute runs (held out $8$ per slice), every law refit only on the computationally inexpensive runs. $R^2$ is the per-group mean; MAPE (\%) is mean\,$\pm$\,std over the slices in each group. \emph{Per-ratio} is a single power law fit independently within each slice. Best per group in \textbf{bold}.}
\label{tab:compute_extrap}
\begin{tabular}{l cc @{\hspace{0.6em}} cc @{\hspace{0.6em}} cc @{\hspace{0.6em}} cc}
\toprule
 & \multicolumn{2}{c}{Undertrained} & \multicolumn{2}{c}{Optimal} & \multicolumn{2}{c}{Overtrained} & \multicolumn{2}{c}{All setups} \\
\cmidrule(lr){2-3}\cmidrule(lr){4-5}\cmidrule(lr){6-7}\cmidrule(lr){8-9}
Law & $R^2$ & MAPE & $R^2$ & MAPE & $R^2$ & MAPE & $R^2$ & MAPE \\
\midrule
Power law (per-ratio) & $0.94$ & $1.32{\pm}0.87$ & $0.96$ & $\mathbf{0.77{\pm}0.39}$ & $0.97$ & $0.86{\pm}0.75$ & $0.95$ & $0.99{\pm}0.69$ \\
Chinchilla            & $0.89$ & $1.52{\pm}0.80$ & $0.47$ & $3.47{\pm}0.65$ & $0.87$ & $1.94{\pm}0.66$ & $0.74$ & $2.34{\pm}1.11$ \\
Farseer               & $\mathbf{0.99}$ & $0.46{\pm}0.20$ & $0.94$ & $1.20{\pm}0.20$ & $\mathbf{0.99}$ & $0.73{\pm}0.23$ & $0.97$ & $0.80{\pm}0.38$ \\
\ska{}                & $\mathbf{0.99}$ & $\mathbf{0.45{\pm}0.21}$ & $\mathbf{0.97}$ & $0.88{\pm}0.15$ & $\mathbf{0.99}$ & $\mathbf{0.42{\pm}0.15}$ & $\mathbf{0.98}$ & $\mathbf{0.60{\pm}0.27}$ \\
\bottomrule
\end{tabular}
\end{table}

\section{Related Work}\label{sec:related}

The modern practice of scaling laws rests on a small set of functional forms. Early work established that model loss falls predictably as a power law in scale~\citep{hestness2017deep,rosenfeld2019constructive}, and \citet{kaplan2020scaling} modeled it jointly in model size $N$ and data $D$, treating the two as coupled. \citet{hoffmann2022training} recast the reducible loss as a purely additive sum of independent power laws, $A/N^{\alpha}+B/D^{\beta}$ above an irreducible term $E$, and this additive form became the field's default. It underlies the familiar compute-optimal rule of roughly $20$ tokens per parameter, reaffirmed in a careful replication~\citep{besiroglu2024chinchilla}, as well as the broader effort to predict the loss of large training runs from much smaller ones~\citep{bi2024deepseek}, a compute-extrapolation problem we study in \Cref{sec:isoratio}. The additive form's convenience, however, conceals a strong assumption: that $N$ and $D$ act on the loss independently, an identically zero mixed derivative $\partial^{2}L/\partial N\,\partial D$, so the marginal value of parameters does not depend on how much data they see.

Efforts to reconcile competing scaling results have targeted the fitting, not the form. When Kaplan- and Chinchilla-style studies disagreed on compute allocation, the gap was traced to parameter counting~\citep{pearce2024reconciling} and to last-layer FLOP accounting, warmup, and optimizer tuning~\citep{porian2024resolving}, all artifacts of the fitting procedure, resolved while leaving the additive form intact. Our evidence points elsewhere: with the fit held fixed, the form itself bends the wrong way at the edges of the grid, where $N$ and $D$ are most imbalanced (\Cref{fig:model,fig:cross_deriv}), and a single coupling exponent is enough to straighten it.

Most closely related, \citet{busbridge2025distillation} use the same untied outer-exponent form, $E+(A/N^{\alpha}+B/D^{\beta})^{\gamma}$, as the supervised scaling law that supplies teacher losses and the corresponding supervised student losses inside their distillation scaling law. Their main contribution is the distillation law: it predicts student loss from teacher loss, student size, and distillation tokens, captures the teacher--student capacity gap, and supports compute-optimal allocation between teacher and student. They state in a footnote that leaving the exponents untied improves fit and extrapolation relative to the tied Kaplan and Chinchilla cases, but do not report a controlled comparison of those supervised forms or study the resulting $N$--$D$ interaction. Our work makes that interaction the central question and provides direct evidence through boundary residuals, mixed derivatives, controlled predictive comparisons, and sampling and extrapolation experiments.

Richer functional forms can also capture the interaction, at a higher cost. The Farseer law~\citep{li2025predictable} makes the data exponent and amplitude depend on $N$, introducing an $N$--$D$ interaction through nine fitted parameters; the added flexibility is harder to fit, and in our experiments it is less accurate than \ska{} at the grid boundaries. Scaling laws have also been extended along other axes: to repeated data, mapped to effective token and parameter counts within the additive form~\citep{muennighoff2023scaling}, and to data mixtures, which add domain weights as inputs to the loss~\citep{ye2024datamixing,shukor2026scaling}. \ska{} instead isolates the effect of a single parameter that couples $N$ and $D$, and we expect similar couplings to matter along other scaling axes.

\section{Conclusion}\label{sec:conclusion}

Additive scaling laws inherently assume that model size and data volume act independently. As we have shown, this assumption breaks down exactly where accurate predictions matter most: the imbalanced extremes of the training grid. The \ska{} law resolves this with a single coupling exponent, a natural and straightforward extension that captures the $N$--$D$ interaction already present in the empirical loss surface. By simply allowing these variables to interact, this single parameter eliminates boundary bias and predicts performance more accurately than both the additive baseline and richer nine-parameter formulations. Furthermore, it yields a strictly sub-additive coupling with a closed-form optimal allocation, a result independently verified by our model-free gradient estimates.

Crucially, because this coupled form is anchored by the edges of the parameter space rather than its interior, we do not need to measure the entire surface to reconstruct it. A sparse, L-shaped grid of inexpensive boundary runs is entirely sufficient to recover the full-grid law at a fraction of the computational cost. Reliable scaling prediction therefore no longer requires massive, dense empirical sweeps, dramatically lowering the barrier to principled compute allocation. We expect this fundamental coupling dynamic to apply along other axes of model scaling, presenting a clear direction for future work.

\bibliographystyle{assets/plainnat}
\bibliography{paper}

\clearpage
\newpage
\beginappendix

\section{Differential analyses}\label{app:diff}
\subsection{Numerical gradients}
\label{app:numerical_gradients}

We estimate derivatives of the loss surface directly from the measured runs. At each grid point, we estimate $\partial z/\partial x_j$ using two mesh-free procedures: a local moving least-squares estimator (MLS) and a global Gaussian-process estimator (GP).

\paragraph{Moving least squares (MLS).} The local estimator fits a truncated Taylor expansion of the surface in a neighbourhood of each query point $x_\star$~\citep{lancaster1981surfaces}. To estimate the derivatives at a specific target point $x_\star$, we locally approximate the loss surface using a multivariate polynomial of degree $p$. This approach comes directly from Taylor expansion: if we fit a local polynomial, its coefficients correspond mathematically to the function's derivatives of various orders.

Let $x$ represent the spatial coordinates, and let $\Delta x_i = x_i - x_\star$ denote the centered distance to a neighboring point. By Taylor's theorem, the local loss value $z(x_i)$ can be expanded as:
$$ z(x_i) \approx c_0 + g^\top \Delta x_i + \frac{1}{2} \Delta x_i^\top H \Delta x_i + \dots $$
where the polynomial expansion continues with higher-order basis terms up to degree $p$. In this formulation, $c_0$ is the local constant, the vector $g$ is the gradient $\nabla z(x_\star)$, and $H$ is the Hessian matrix containing our target cross-derivative.

To solve for these unknowns simultaneously, we flatten all the coefficients of this expansion into a single vector $c$. We then determine the optimal coefficient estimate $\hat{c}$ by fitting this polynomial to the $k$ nearest neighbors using distance-weighted ridge regression:
$$ \hat{c} = \big(\Phi^\top W \Phi + \lambda I\big)^{-1}\Phi^\top W z $$

The components of this generalized regression map directly to the Taylor expansion:
\begin{itemize}
    \item $\Phi$ is the design matrix. Each row corresponds to a neighbor and contains its evaluated polynomial basis terms up to degree $p$ (for example: $1$, $\Delta x_{i,1}$, $\Delta x_{i,2}$, $\Delta x_{i,1}^2$, the interaction term $\Delta x_{i,1}\Delta x_{i,2}$, and higher-order combinations). 
    \item $W$ is a diagonal weight matrix where $w_i=\exp\!\big(\!-\|\Delta x_i\|^2/\sigma^2\big)$. This acts as a Gaussian filter, ensuring that points closer to $x_\star$ exert more influence on the local fit. The bandwidth $\sigma$ scales dynamically with the neighborhood radius.
    \item $\lambda I$ adds a small ridge regularization penalty to keep the matrix inversion numerically stable, which is crucial if the scattered local data points are poorly distributed.
\end{itemize}

Once we solve for $\hat{c}$, extracting the necessary derivatives is straightforward. We simply read them directly from their designated slots in the vector: the first-order block provides the gradient estimate, and the coefficient corresponding to the $\Delta x_{i,1}\Delta x_{i,2}$ basis term provides the exact mixed cross-derivative required for our analysis. Because this Moving Least Squares (MLS) approach relies entirely on local geometry, its accuracy is naturally sensitive to local noise, the chosen neighborhood size $k$, and the polynomial degree $p$.

\paragraph{Gaussian process (GP).} The global estimator fits a single Gaussian process to all points and differentiates its posterior mean in closed form. We model the empirical loss targets $z \in \mathbb{R}^{n \times 1}$ as a smooth function of the input coordinates, corrupted by some observation noise $\sigma_n^2$. In standard GP regression, the predicted log-loss $\hat{z}(x_\star)$ at a target point $x_\star$ is given directly by the posterior mean:
$$ \hat{z}(x_\star) = \mu(x_\star) = k(x_\star, X)\alpha $$
Here, $k(x_\star, X) \in \mathbb{R}^{1 \times n}$ is a row vector of kernel evaluations between the target point and all $n$ training data points. The learned weight vector is defined as $\alpha = (K + \sigma_n^2 I)^{-1}z \in \mathbb{R}^{n \times 1}$, where $K \in \mathbb{R}^{n \times n}$ is the dense covariance matrix of the training inputs and $I$ is the $n \times n$ identity matrix.

The key advantage of using a GP is that differentiation is a linear operator, meaning the derivative of a GP is simply another GP. We can compute the estimated scalar gradient of the loss surface in closed form by directly differentiating the predicted surface $\hat{z}(x_\star)$ with respect to the input dimensions. Because the weights $\alpha$ are independent of $x_\star$, the derivative simply applies to the $1 \times n$ row vector:
$$ \widehat{\frac{\partial z}{\partial x_j}}(x_\star) = \frac{\partial \hat{z}(x_\star)}{\partial x_j} = \frac{\partial k(x_\star,X)}{\partial x_j} \alpha $$

Because standard kernel functions are analytically differentiable, this computation is exact. In this work, we are using a Radial Basis Function (RBF) kernel with length-scale $\ell_j$, the derivative is $\partial k(x,x')/\partial x_j = k(x,x')(x'_j-x_j)/\ell_j^2$. Composite kernels can be differentiated using standard sum and product rules, provided that any additive white-noise terms are excluded from the predictive kernel.

Additionally, the GP provides a closed-form predictive variance for this gradient, quantifying our estimation uncertainty. It is calculated by taking the scalar prior variance of the gradient and subtracting the information gained from the observed data:
$$ \mathrm{Var}\!\left[\frac{\partial z}{\partial x_j}(x_\star)\right] = \left.\frac{\partial^2 k(x,x')}{\partial x_j\,\partial x'_j}\right|_{x=x'=x_\star} - \frac{\partial k(x_\star,X)}{\partial x_j}\,(K+\sigma_n^2 I)^{-1}\,\frac{\partial k(X,x_\star)}{\partial x_j} $$
Notice how the matrix dimensions elegantly resolve to a scalar: a $1 \times n$ vector multiplied by an $n \times n$ matrix, multiplied by an $n \times 1$ column vector $\frac{\partial k(X,x_\star)}{\partial x_j}$.

Unlike MLS, which requires manual tuning of neighborhood sizes and polynomial degrees, the GP explicitly models observation noise and automatically tunes its length-scales and noise levels by maximizing the marginal likelihood.

\subsection{Form of the coupling}
\label{app:cross_deriv}
The non-zero mixed derivative of \Cref{sec:motivation} constrains the functional form: it should preserve the marginal power-law behaviour visible in the first derivatives without making the two axes additively separable. The \ska{} form (\Cref{eq:skaling}) satisfies this. Writing $u=A N^{-\alpha}+B D^{-\beta}$, its mixed derivative is
\[
\frac{\partial^{2}L}{\partial N\,\partial D}
=
k(k-1)\,u^{k-2}\,
\alpha A\,N^{-\alpha-1}\,
\beta B\,D^{-\beta-1}.
\]
Thus $k=1$ recovers the additive Chinchilla law and forces the mixed derivative to vanish, while any $k\neq1$ introduces an interaction. In the empirically relevant case $0<k<1$, the mixed derivative is negative, matching the sign observed in \Cref{fig:cross_deriv}. At the same time, the first derivatives,
\[
\frac{\partial L}{\partial N}
=
-k\alpha A\,N^{-\alpha-1}u^{k-1},
\qquad
\frac{\partial L}{\partial D}
=
-k\beta B\,D^{-\beta-1}u^{k-1},
\]
remain dominated by the same-variable power-law factors, with cross-variable dependence entering only through the shared factor $u^{k-1}$. Consequently, the \ska{} law can appear nearly separable at first order while still allowing the non-zero mixed derivative indicated by the data.

\paragraph{Why a multiplicative coupling, not an additive interaction term.} A natural alternative to the \ska{} law would be to keep the Chinchilla law additive and append a separate product term,
\[
L = A N^{-\alpha} + B D^{-\beta} + G\,N^{-\mu}D^{-\nu} + E .
\]
This term can create a non-zero mixed derivative, but its sign creates an immediate constraint. For this model, the mixed derivative and the contribution of the interaction term to the size derivative are
\[
\frac{\partial^2 L}{\partial N\,\partial D}
=\mu\nu\,G\,N^{-\mu-1}D^{-\nu-1},
\qquad
\frac{\partial L}{\partial N}\!\left(GN^{-\mu}D^{-\nu}\right)
=-\,\mu\,G\,N^{-\mu-1}D^{-\nu}.
\]
Thus the sign of $G$ controls two quantities in opposite ways. Matching the observed negative mixed derivative requires $G<0$. But with $G<0$, the interaction term contributes positively to $\partial L/\partial N$, opposing the desired monotonic decrease with model size. This is not just a local inconvenience: the fitted mixed-derivative decay implies $\mu\approx0.1$, smaller than the main size exponent $\alpha$, so this positive contribution decays more slowly in $N$ than the leading negative term. At large $N$ and small $D$, it can therefore dominate and make $\partial L/\partial N>0$, predicting that loss increases when the model becomes larger. Choosing $G>0$ avoids this monotonicity failure, but then the mixed derivative becomes positive and the desired synergy disappears. Thus a single additive product term cannot simultaneously preserve monotonicity and match the observed negative interaction.

\ska{} avoids this sign conflict because the interaction is introduced by a positive multiplicative factor rather than by a separate signed term. Its size gradient is
\[
\frac{\partial L}{\partial N}=-k\,\alpha A\,N^{-\alpha-1}u^{k-1},
\]
which is negative for every $k>0$. The dependence on data enters only through the positive factor $u^{k-1}$. When $0<k<1$, increasing $D$ decreases $u$, which increases $u^{k-1}$ and therefore increases the magnitude of the already-negative size gradient. This yields $\partial^2 L/\partial N\partial D<0$ without ever changing the sign of $\partial L/\partial N$. In this parameterization, monotonicity and synergy are compatible by construction.

The same factor also explains the asymmetric first-order slopes in \Cref{fig:deriv_deps}. Let $w_D=BD^{-\beta}/u$ and $w_N=AN^{-\alpha}/u$ denote the data and size shares of the inner sum. Differentiating $\ln|\partial L/\partial N|$ and $\ln|\partial L/\partial D|$ gives $\gamma_N=(1-k)\,\beta\,w_D$ and $\gamma_D=(1-k)\,\alpha\,w_N$, so $\gamma_N/\gamma_D=(\beta/\alpha)\,(w_D/w_N)$. This ratio exceeds one whenever $\beta>\alpha$ and the inner sum is data-leaning, reproducing the measured $\gamma_N\approx0.13>\gamma_D\approx0.07$ through the single coupling exponent rather than through a skewed interaction term.

\subsection{Empirical optimal token to parameter ratio}
\label{app:emp_opt_ratio}

The numerical gradients also provide an empirical estimate of the compute-optimal allocation, without fitting a parametric scaling law. For a fixed training budget $C=6ND$, feasible configurations lie on a single iso-compute curve: increasing model size requires decreasing the number of training tokens, and conversely. The empirical optimum for that budget is the lowest-loss point on this curve,
\[
\min_{N,D}\; L(N,D) \quad \mathrm{s.t.} \quad 6ND = C .
\]

For an interior optimum on a differentiable loss surface, the Lagrangian
\[
\mathcal{J}(N,D,\lambda)=L(N,D)+\lambda(6ND-C)
\]
has stationarity conditions
\[
\frac{\partial L}{\partial N} = -6\lambda D, \qquad \frac{\partial L}{\partial D} = -6\lambda N .
\]
Dividing the two equations cancels the multiplier and leaves a condition on the loss gradients alone:
\begin{equation}
N\,\frac{\partial L}{\partial N} = D\,\frac{\partial L}{\partial D}
\quad\Longleftrightarrow\quad
\frac{\partial \ln L}{\partial \ln N} = \frac{\partial \ln L}{\partial \ln D}.
\end{equation}
Thus, at a compute-optimal point, a $1\%$ increase in model size and a $1\%$ increase in data have the same signed effect on the loss. Equivalently, the directional derivative along the tangent to the iso-compute curve must vanish. This is a constrained stationarity condition: the full gradient is not required to vanish, since increasing the total compute would still reduce loss.

To solve this condition non-parametrically, we write the loss in log coordinates,
\[
x=\ln N,\qquad y=\ln D,\qquad z(x,y)=\ln L(N,D).
\]
The optimality condition becomes $\partial z/\partial x=\partial z/\partial y$. We locate this equality using the numerical log-slopes from \Cref{app:numerical_gradients}; the GP and MLS estimators give two fully independent estimates of the frontier through the same construction.

\paragraph{GP surrogate.} The optimal compute allocation occurs exactly where the partial derivatives balance. Because our empirical data consists of discrete points, we cannot locate this exact equilibrium directly from the grid. Instead, we use the GP posterior mean to provide a continuous, differentiable surface. We can then simply track the points on this surface where the difference between the two gradients is exactly zero. This continuous zero-crossing boundary directly yields the optimal token-to-parameter ratio for any compute budget. If multiple roots exist for a given budget, we selects the minimum with the lowest predicted loss.

\paragraph{MLS estimates.} The local estimator is itself a mesh-free surrogate, so we follow exactly the same procedure: around any query point it fits a weighted polynomial to the $k$ nearest runs, from which we read the gradient difference on a gride of point track its zero-crossing as above.

\begin{figure}[!b]
\centering
\includegraphics[width=\linewidth]{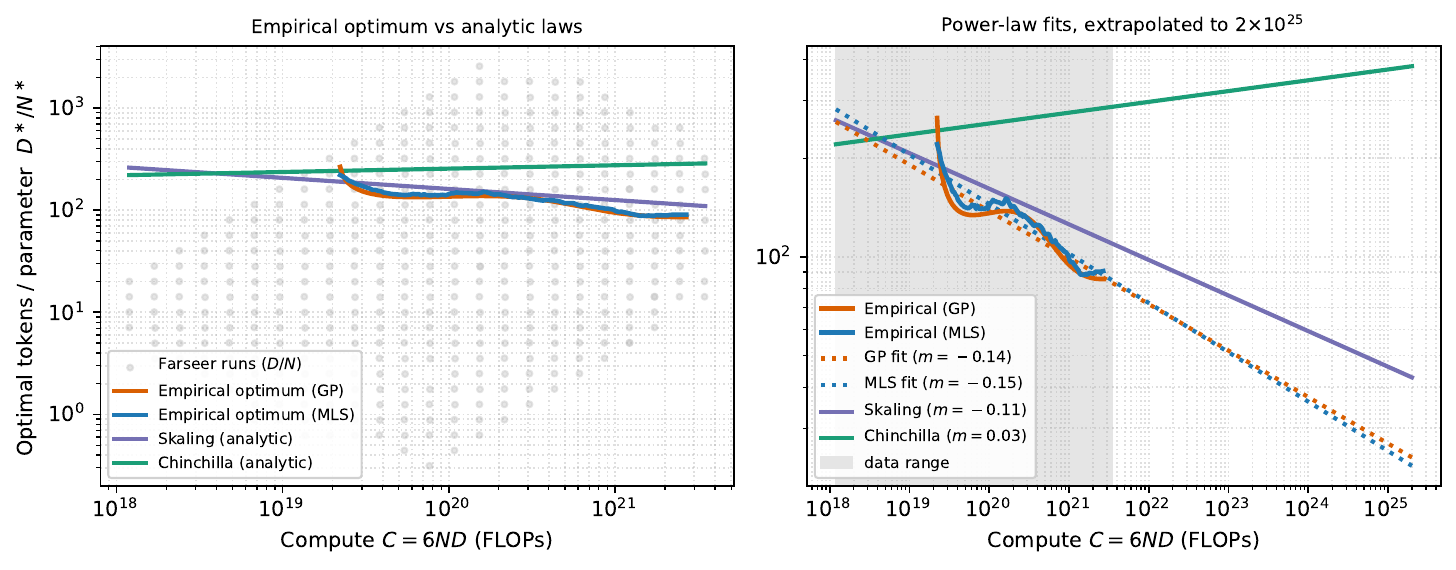}
\caption{\textbf{Empirical compute-optimal token-to-parameter ratio on Farseer, recovered without a parametric fit.} The optimum $D^\star/N^\star$ is located where the log-slopes balance ($\partial\ln L/\partial\ln N=\partial\ln L/\partial\ln D$), using a global GP surrogate and a local MLS surrogate (both evaluated as continuous mesh-free fields), and compared with the closed-form $R_{opt}$ of \ska{} and Chinchilla fitted on the same $(N,D)$ data. \emph{Left:} both empirical optima, over the run cloud, follow \ska{} more closely than the nearly flat Chinchilla prediction. \emph{Right:} power-law fits of each empirical frontier (dotted) against the analytic laws, extrapolated to $2\times10^{25}$ FLOPs (shaded $=$ data range); the empirical exponents ($-0.14$, $-0.15$) are close to \ska{} ($-0.11$) and have the opposite sign from Chinchilla ($+0.03$).}
\label{fig:opt_ratio}
\end{figure}

In both cases, tracking $D^\star/N^\star$ as $C$ varies gives an empirical token-to-parameter ratio, which we compare with the closed-form $R_{opt}$ of \Cref{app:compute_optimal} on the Farseer data (\Cref{fig:opt_ratio}). The two independent estimates agree qualitatively: both recover a ratio that decreases with compute, closely following \ska{}'s $R_{opt}$, whereas Chinchilla predicts a nearly flat ratio. Fitting each empirical frontier as a power law $D^\star/N^\star\propto C^{m}$ gives $m\approx-0.14$ for the GP and $m\approx-0.15$ for MLS, close to \ska{}'s exponent of $-0.11$ and opposite in sign to Chinchilla's $+0.03$. When extrapolated one order of magnitude beyond the data, to $2\times10^{25}$ FLOPs, the allocations differ by more than $10\times$: Chinchilla approaches $\sim\!380$ tokens per parameter, while the empirical fits and the \ska{} law fall to $20$--$40$. The coupling therefore has a material effect on large-scale allocation.

\section{Compute-optimal training}
\label{app:compute_optimal}

The primary utility of a scaling law is to determine the optimal allocation of a given compute budget $C$ between model size $N$ and data volume $D$. We formally define the training compute budget as $C = 6ND$. To identify the compute-optimal configuration, we minimize the \ska{} loss of \Cref{eq:skaling} subject to this budget constraint.

By substituting $D = \frac{C}{6N}$, we can isolate the inner additive term as a function of $N$, defining $Z(N) = \frac{A}{N^{\alpha}} + B \left(\frac{C}{6N}\right)^{-\beta}$. Our loss objective then simplifies to:
\begin{equation}
    L(N) = Z(N)^{k} + E
\end{equation}

Taking the derivative with respect to $N$ yields:
\begin{equation}
    \frac{dL}{dN} = k \cdot Z(N)^{k-1} \cdot Z'(N)
\end{equation}

Since $Z(N)$ is a sum of strictly positive terms and our empirical fits consistently yield $k > 0$, the scaling factor $k \cdot Z(N)^{k-1}$ is non-zero. This means minimizing the loss strictly requires that $Z'(N) = 0$, precisely the stationarity condition of the additive Chinchilla law~\citep{hoffmann2022training}; the \ska{} law therefore inherits Chinchilla's compute-optimal allocation unchanged. Solving $Z'(N) = -\alpha A\,N^{-\alpha-1} + \beta B\,(6/C)^{\beta} N^{\beta-1} = 0$ gives $N^{\alpha+\beta} = \frac{\alpha A}{\beta B}(C/6)^{\beta}$; substituting $D^{*} = C/(6N^{*})$ then yields the optimal token-to-parameter ratio (i.e., $D^*/N^*$):
\begin{equation}
    R_{opt} = 6^{\frac{\beta-\alpha}{\alpha+\beta}} \left(\frac{\beta B}{\alpha A}\right)^{\frac{2}{\alpha+\beta}} C^{\frac{\alpha-\beta}{\alpha+\beta}}
\end{equation}

This reveals that if $\alpha \approx \beta$ (as observed in Chinchilla), the optimal ratio $D/N$ remains constant across scales. More broadly, the \ska{} law cleanly separates the shape of the loss landscape (controlled by $k$) from the location of the optimal allocation ($Z'(N) = 0$): the monotone outer map $x \mapsto x^{k} + E$ rescales the loss but leaves its minimizer unchanged, so the \ska{} law gains expressivity without sacrificing tractability.

\section{Fitting Details}
\label{app:fitting}
\subsection{Challenges}
Despite the small number of parameters, fitting scaling laws to empirical loss data is a surprisingly fragile non-convex optimization problem. Several structural factors contribute to this instability. First, the parameters operate on vastly different numerical scales. For instance, the large magnitudes of coefficients $A$ and $B$ contrast sharply with the fractional exponents $\alpha$ and $\beta$, ill-conditioning the problem for standard gradient-based optimizers like L-BFGS. Second, the optimization landscape is notoriously flat. Because the loss decreases logarithmically, optimizers are prone to halting prematurely, a vulnerability that is worsened when relying on finite-difference gradient approximations. Finally, the parameters exhibit strong compensatory behaviors; the irreducible loss $E$, in particular, is difficult to accurately estimate because other coefficients can easily shift to offset its value. This creates wide valleys of distinct local minima that yield nearly identical overall surface fits.

We fit every law by minimizing the objective in log space, which (as in Chinchilla) absorbs much of this scale disparity. Within this setup, a carefully configured L-BFGS, paired with basin-hopping and automatic differentiation~\citep{besiroglu2024chinchilla,shukor2026scaling}, and the scale-independent, gradient-free CMA-ES reach equally good fits. The difference we observe is practical rather than in final quality: L-BFGS must be carefully tuned (initialization, restarts, and the log-space objective) to reach them reliably, whereas CMA-ES attains the same solutions out of the box, without any such tweaks.

\subsection{Hyperparameters}

We implement two global optimization strategies. The first, L-BFGS-B with basin-hopping, pairs a gradient-based local optimizer (L-BFGS-B with analytical gradients via autograd) with multi-start basin-hopping for global search; starting points are drawn from a Sobol quasi-random sequence over the bounded parameter space. The second, BIPOP-CMA-ES~\citep{hansen2016cma}, is a gradient-free evolutionary strategy; we use doubled population size, active CMA, and 9 BIPOP restarts. As discussed in \Cref{app:fitting}, both reach equally good fits; all results reported in this paper use L-BFGS-B with basin-hopping, as it is more common in the literature~\citep{besiroglu2024chinchilla,shukor2026scaling}. Every law minimizes a Huber loss ($\delta=0.05$) in log space; coefficients $A$ and $B$ are optimized in log scale to absorb their large dynamic range. The per-law configurations are summarized in \Cref{tab:fit_hparams}.

\begin{table}[h]
\centering
\footnotesize
\setlength{\tabcolsep}{4pt}
\caption{Per-law parameter bounds and fitting configuration. All laws use $2000$ basin-hopping restarts, a log-space Huber objective, and analytic (autograd) gradients; \emph{Params} is the number of free parameters and \emph{Log params} those optimized in log scale. Farseer uses the ExpExpExp parameterization of \citep{li2025predictable}.}
\label{tab:fit_hparams}
\begin{tabular}{l c l p{0.56\linewidth}}
\toprule
Law & Params & Log params & Bounds \\
\midrule
Chinchilla & 5 & $A,B$ & $A\!\in\![10^{-6},10^{4}]$, $B\!\in\![10^{-6},5{\times}10^{4}]$, $\alpha,\beta\!\in\![0,1]$, $E\!\in\![0,3]$ \\
\ska{}     & 6 & $A,B$ & $A,B\!\in\![10^{-6},10^{7}]$, $\alpha,\beta,k\!\in\![0.01,2]$, $E\!\in\![0,3]$ \\
Farseer    & 9 & ---   & $E\!\in\![0gr@,5]$, $s\!\in\![-10,10]$, $q\!\in\![-0.5,0.5]$, $S\!\in\![-30,10]$, $B_c\!\in\![-10,500]$, $b\!\in\![-2,0.5]$, $Q\!\in\![-10,15]$, $A_c\!\in\![-25,5]$, $a\!\in\![-0.5,0.2]$ \\
\bottomrule
\end{tabular}
\end{table}

\paragraph{Fitting Farseer.} We fit Farseer, like every other law, with this common direct optimizer, so the comparison reflects the functional form and not the fitting recipe. This is also a practical necessity: Farseer's original pipeline estimates its components from consecutive-$D$ differences at fixed $N$, which requires several evenly spaced $D$ values per model size. The sparse L-shape grids and the scattered, non-gridded datasets do not provide this, and even on a fully gridded dataset the cross-validation hold-outs remove points and break the difference structure the pipeline relies on; direct optimization is the only procedure that applies uniformly across our settings. Run under the same cross-validation, the original pipeline did not give significantly better predictions: its errors lay within the fold-to-fold standard deviation of the direct fit in every regime except larger-$D$ extrapolation, where it only matched \ska{}. We were unable to fit this form substantially better with any of these procedures, so we report the common direct fit as the best Farseer result we obtained.

\section{Additional datasets}
\label{app:extra_datasets}
We repeat the cross-validation protocol of \Cref{sec:evaluation_protocol} on two further datasets (\Cref{tab:cv_mape_extra}). \emph{Farseer-code} is the code-domain counterpart of the Farseer grid~\citep{li2025predictable}: $117$ runs over $9$ model sizes ($201$M--$3.18$B) and $20$ token budgets ($2$B--$128$B), spanning compute $C{=}6ND$ from $2.4\times10^{18}$ to $2.4\times10^{21}$ FLOPs. Being grid-structured, it admits both the full-grid and L-shape splits; its larger-$N$ and larger-$D$ hold-outs are the largest model sizes and longest token horizons at the top edge of the grid. The \emph{Chinchilla} loss measurements~\citep{besiroglu2024chinchilla} are $245$ scattered $(N,D)$ points covering $57$M--$16.2$B parameters and $245$M--$318$B tokens ($1.4\times10^{18}$--$1.3\times10^{22}$ FLOPs); they do not lie on a regular grid, so only the full-grid (random) split applies (no L-shape), with the largest-$N$ ($16.2$B) and largest-$D$ ($318$B-token) points forming the two extrapolation sets.

\begin{table}[t]
\centering
\footnotesize
\setlength{\tabcolsep}{2pt}
\caption{Fit quality and predictive error on two additional datasets, following the protocol of \Cref{tab:cv_mape}: interpolation $R^2$ ($\uparrow$) and MAPE (\%, $\downarrow$; mean\,$\pm$\,std over 5 CV folds) on interpolation and two extrapolation regimes (larger $N$, larger $D$). Farseer-code is grid-structured (full and L-shape splits); the Chinchilla measurements are not gridded, so only a full-grid split applies. Best per column within each block in \textbf{bold}.}
\label{tab:cv_mape_extra}
\resizebox{\linewidth}{!}{%
\begin{tabular}{l cccc @{\hspace{0.5em}} cccc @{\hspace{0.5em}} cccc}
\toprule
 & \multicolumn{4}{c}{Farseer-code (full)} & \multicolumn{4}{c}{Farseer-code (L-shape)} & \multicolumn{4}{c}{Chinchilla (full)} \\
\cmidrule(lr){2-5} \cmidrule(lr){6-9} \cmidrule(lr){10-13}
Law & $R^2$ & Interp. & Ext.\,$N$ & Ext.\,$D$ & $R^2$ & Interp. & Ext.\,$N$ & Ext.\,$D$ & $R^2$ & Interp. & Ext.\,$N$ & Ext.\,$D$ \\
\midrule
Chinchilla            & $0.998$ & $0.28{\pm}0.07$ & $0.93{\pm}0.15$ & $0.60{\pm}0.04$ & $0.991$ & $0.59{\pm}0.06$ & $1.93{\pm}0.18$ & $1.60{\pm}0.12$ & $\mathbf{0.993}$ & $0.63{\pm}0.15$ & $\mathbf{1.16{\pm}0.33}$ & $0.63{\pm}0.06$ \\
Farseer               & $0.983$ & $0.89{\pm}0.12$ & $1.01{\pm}0.04$ & $3.74{\pm}0.23$ & $0.993$ & $0.53{\pm}0.16$ & $\mathbf{0.96{\pm}0.27}$ & $\mathbf{0.91{\pm}0.68}$ & $0.955$ & $1.27{\pm}0.55$ & $1.23{\pm}1.10$ & $1.12{\pm}0.21$ \\
\ska{}                & $\mathbf{0.999}$ & $\mathbf{0.24{\pm}0.09}$ & $\mathbf{0.67{\pm}0.19}$ & $\mathbf{0.26{\pm}0.06}$ & $\mathbf{0.995}$ & $\mathbf{0.45{\pm}0.12}$ & $1.39{\pm}0.49$ & $1.14{\pm}0.41$ & $\mathbf{0.993}$ & $\mathbf{0.61{\pm}0.14}$ & $1.28{\pm}0.29$ & $\mathbf{0.51{\pm}0.03}$ \\
\bottomrule
\end{tabular}}
\end{table}

\Cref{tab:params_extra} lists the corresponding coefficients for Farseer-code and the Chinchilla data. The same pattern holds, with the \ska{} law fitting $k<1$ and a smaller $E$ than Chinchilla, though the coupling is weaker here ($k\approx0.77$--$0.90$), in line with the more mixed accuracy gains in \Cref{tab:cv_mape_extra}.

\begin{table}[t]
\centering
\footnotesize
\setlength{\tabcolsep}{4pt}
\caption{Fitted coefficients (mean\,$\pm$\,std over $5$ CV folds) behind \Cref{tab:cv_mape_extra}. Chinchilla is the $k{=}1$ special case of \ska{}.}
\label{tab:params_extra}
\resizebox{\linewidth}{!}{%
\begin{tabular}{ll cccccc}
\toprule
Setup & Law & $A$ & $B$ & $\alpha$ & $\beta$ & $k$ & $E$ \\
\midrule
Farseer-code, full & Chinchilla & $(1.8{\pm}1.1){\times}10^{3}$ & $(5.6{\pm}0.8){\times}10^{2}$ & $0.48{\pm}0.03$ & $0.34{\pm}0.01$ & $1$ & $0.65{\pm}0.01$ \\
 & \ska{} & $(4.2{\pm}2.8){\times}10^{3}$ & $(2.0{\pm}0.2){\times}10^{3}$ & $0.52{\pm}0.03$ & $0.40{\pm}0.01$ & $0.77{\pm}0.03$ & $0.60{\pm}0.02$ \\
\addlinespace
Farseer-code, L-shape & Chinchilla & $(5.4{\pm}1.2){\times}10^{3}$ & $(1.5{\pm}0.1){\times}10^{3}$ & $0.55{\pm}0.01$ & $0.39{\pm}0.00$ & $1$ & $0.71{\pm}0.00$ \\
 & \ska{} & $(4.1{\pm}1.4){\times}10^{3}$ & $(2.1{\pm}0.5){\times}10^{3}$ & $0.53{\pm}0.02$ & $0.40{\pm}0.01$ & $0.90{\pm}0.06$ & $0.67{\pm}0.03$ \\
\addlinespace
Chinchilla, full & Chinchilla & $(7.0{\pm}1.5){\times}10^{2}$ & $(1.3{\pm}0.3){\times}10^{4}$ & $0.37{\pm}0.02$ & $0.45{\pm}0.01$ & $1$ & $1.91{\pm}0.02$ \\
 & \ska{} & $(4.9{\pm}3.0){\times}10^{3}$ & $(1.1{\pm}0.7){\times}10^{5}$ & $0.45{\pm}0.04$ & $0.53{\pm}0.03$ & $0.77{\pm}0.06$ & $1.85{\pm}0.01$ \\
\bottomrule
\end{tabular}}
\end{table}

\section{Pre-training Details}
\label{app:pretraining}
\subsection{Model}
The configurations in this section describe the SK-Grid, an internal set of pretraining runs; for the Farseer grid we use the released runs and refer the reader to \citep{li2025predictable} for details. SK-Grid comprises $125$ runs across $14$ model sizes (\Cref{tab:model_configs}).

\paragraph{Architecture.} The models span $14$ sizes from $134$M to $4.9$B parameters, scaled by growing width and depth together: the model dimension $d_{\text{model}}$ increases from $672$ to $3264$ and the depth from $7$ to $34$ layers. All models use the Llama~3 tokenizer, with a vocabulary of $128{,}256$.

\paragraph{Token budgets.} Each size is trained on a geometric ladder of token budgets spanning $316$M to $316$B tokens, five budgets per decade. Per-run compute is capped, so larger models are trained on fewer budgets: the smallest models cover all $16$ horizons while the largest is trained on a single budget, producing the staircase grid whose row counts appear in the last column of \Cref{tab:model_configs}.

\subsection{Hyperparameters}
The two run-specific hyperparameters, the global batch size $B$ (tokens) and the peak learning rate $\eta$, follow the StepLaw prescription~\citep{li2025hyperopt} as power laws in the per-token compute $F$ (FLOPs per token, a monotone proxy for model size $N$) and the token budget $D$, with coefficients refit to our own setup:
\begin{equation}\label{eq:hp}
B = 896.07\,F^{0.231},
\qquad
\eta = 0.0709\,F^{-0.4303}\,D^{0.2785}.
\end{equation}
Because the shape of the loss surface is highly sensitive to these tuning choices, different parameterizations of hyperparameters in the training recipe can alter the apparent $N$--$D$ interaction and shift the inferred compute-optimal ratio. For instance, a broadly mistuned grid might artificially dampen the measured interaction or skew the optimal ratio. This sensitivity also implies that differences in hyperparameter policies, such as the original Chinchilla recipe compared to our StepLaw-tuned SK-Grid, partly explain why different datasets exhibit varying degrees of $N$--$D$ coupling. Consequently, cross-dataset comparisons inherently reflect the specific training recipes aand data used to generate each grid.

All remaining settings are fixed across the grid (\Cref{tab:hparams}), based on the defaults of the Meta Lingua framework~\citep{meta_lingua}. Every run uses the same data mixture: $60\%$ DCLM-Edu web text~\citep{li2024datacomp}, $30\%$ code, and $10\%$ math. The loss we fit is the validation loss on a held-out split of this mixture. For the other datasets, refer to their respective papers: Farseer~\citep{li2025predictable} and Chinchilla~\citep{besiroglu2024chinchilla,hoffmann2022training}.

\begin{table}[ht]
\centering
\scriptsize
\begin{minipage}[t]{0.5\linewidth}
\centering
\setlength{\tabcolsep}{3pt}
\caption{SK-Grid model configurations.}
\label{tab:model_configs}
\begin{tabular}{rccccc}
\toprule
$N$ & $d_{\text{model}}$ & Layers & Heads & $d_{\text{head}}$ & Budgets \\
\midrule
$134$M  & $672$  & $7$  & $14$ & $48$ & $16$ \\
$177$M  & $864$  & $9$  & $9$  & $96$ & $16$ \\
$234$M  & $960$  & $10$ & $10$ & $96$ & $15$ \\
$308$M  & $1056$ & $11$ & $11$ & $96$ & $14$ \\
$407$M  & $1248$ & $13$ & $13$ & $96$ & $11$ \\
$537$M  & $1440$ & $15$ & $15$ & $96$ & $11$ \\
$708$M  & $1632$ & $17$ & $17$ & $96$ & $9$  \\
$935$M  & $1824$ & $19$ & $19$ & $96$ & $9$  \\
$1.23$B & $2016$ & $21$ & $21$ & $96$ & $7$  \\
$1.63$B & $2208$ & $23$ & $23$ & $96$ & $5$  \\
$2.15$B & $2496$ & $26$ & $26$ & $96$ & $5$  \\
$2.83$B & $2688$ & $28$ & $28$ & $96$ & $4$  \\
$3.74$B & $2976$ & $31$ & $31$ & $96$ & $2$  \\
$4.93$B & $3264$ & $34$ & $34$ & $96$ & $1$  \\
\bottomrule
\end{tabular}
\end{minipage}\hfill
\begin{minipage}[t]{0.48\linewidth}
\centering
\setlength{\tabcolsep}{5pt}
\caption{Fixed configuration shared by all SK-Grid runs; per-run LR and batch size follow \Cref{eq:hp}.}
\label{tab:hparams}
\begin{tabular}{ll}
\toprule
\multicolumn{2}{l}{\emph{Architecture}} \\
Position enc. & RoPE ($\theta{=}10^{4}$) \\
Vocabulary    & $128{,}256$ \\
Sequence len. & $2048$ \\
\addlinespace
\multicolumn{2}{l}{\emph{Optimization}} \\
Optimizer     & AdamW \\
$(\beta_1,\beta_2)$ & $(0.9,\,0.95)$ \\
Weight decay  & $0.1$ \\
Grad. clip    & $0.1$ \\
LR schedule   & cosine \\
Warmup        & $10\%$ \\
Final LR      & $1 \times 10^{-6}$ \\
\bottomrule
\end{tabular}
\end{minipage}
\end{table}

\section{Dominated-pair fitting}
\label{app:dompair}
A recurring difficulty above is that the irreducible loss $E$ is only weakly identified: interior points constrain it poorly and the other coefficients shift to absorb it. The \emph{dominated-pair} fit removes $E$ from the objective entirely. For any ordered pair in which configuration $i$ dominates $j$ ($F_i\ge F_j$, $D_i\ge D_j$ and $L_i<L_j$), the additive floor cancels in the loss difference:
\begin{equation}\label{eq:dompair}
L_j-L_i=\big(A\,F_j^{-\alpha}+B\,D_j^{-\beta}\big)^{k}-\big(A\,F_i^{-\alpha}+B\,D_i^{-\beta}\big)^{k},
\end{equation}
which no longer involves $E$ (Chinchilla is the $k{=}1$ case). We fit the shape parameters $(A,B,\alpha,\beta,k)$ on all such pairwise differences, then recover the floor as $E=\operatorname{median}_k\big(L_k-(A\,F_k^{-\alpha}+B\,D_k^{-\beta})^{k}\big)$, decoupling the reducible shape from the constant offset.

\Cref{tab:brute} shows that dominated-pair fitting serves primarily as a correction for the additive Chinchilla law. Across both datasets, it reduces most of Chinchilla's extrapolation errors, yielding the clearest gains in boundary regimes where accurately fitting the irreducible loss floor matters most. For example, on the full grids, Chinchilla's far-extrapolation error drops from $2.46\%$ to $0.79\%$ on Farseer and from $5.17\%$ to $3.67\%$ on SK-Grid. The L-shape grids show similar improvements across all reported metrics. 

This consistent pattern suggests that much of Chinchilla's extrapolation error stems from weak identification of the floor $E$; once $E$ is removed from the objective, the constant offset no longer distorts the fit of the reducible component. Furthermore, because Chinchilla's functional form does not perfectly capture the global shape of the loss surface, the dominated-pair objective implicitly reweights the data, forcing the optimization to anchor more heavily on the extreme boundary points of the grid. In contrast, this correction does not consistently improve \ska{}.

\begin{table}[t]
\centering
\footnotesize
\setlength{\tabcolsep}{2pt}
\caption{Dominated-pair (``+dom'') fitting versus the default joint L-BFGS fit, for Chinchilla and \ska{} on every dataset and grid. Each ``+dom'' row is the dominated-pair fit (\Cref{eq:dompair}) of the law above it. Columns: interpolation $R^2$ ($\uparrow$) and MAPE (\%, $\downarrow$; mean\,$\pm$\,std over 5 CV folds) on interpolation and the larger-$N$, larger-$D$, and far regimes. The Chinchilla measurements are not gridded, so they have no L-shape and no far set; the Farseer-code far set is a single held-out run and is omitted (\,--\,). Both variants come from one fitting run, so the baselines can differ marginally from \Cref{tab:cv_mape}. Best per column within each panel in \textbf{bold} (far excluded where not comparable).}
\label{tab:brute}
\begin{tabular}{l ccccc @{\hspace{0.5em}} ccccc}
\toprule
Law & $R^2$ & Interp. & Ext.\,$N$ & Ext.\,$D$ & Far & $R^2$ & Interp. & Ext.\,$N$ & Ext.\,$D$ & Far \\
\midrule
 & \multicolumn{5}{c}{\textit{Farseer (full)}} & \multicolumn{5}{c}{\textit{Farseer (L-shape)}} \\
\cmidrule(lr){2-6}\cmidrule(lr){7-11}
Chinchilla & $0.995$ & $0.77{\pm}0.04$ & $1.48{\pm}0.03$ & $1.98{\pm}0.08$ & $2.46{\pm}0.19$ & $0.954$ & $2.51{\pm}0.07$ & $4.32{\pm}0.13$ & $3.29{\pm}0.11$ & $9.82{\pm}0.48$ \\
\quad +dom & $0.995$ & $0.82{\pm}0.06$ & $1.40{\pm}0.06$ & $2.61{\pm}0.22$ & $\mathbf{0.79{\pm}0.08}$ & $0.959$ & $2.41{\pm}0.17$ & $3.82{\pm}0.18$ & $2.89{\pm}0.13$ & $7.84{\pm}0.87$ \\
\ska{} & $\mathbf{0.998}$ & $\mathbf{0.41{\pm}0.05}$ & $0.47{\pm}0.03$ & $0.88{\pm}0.06$ & $2.32{\pm}0.18$ & $0.995$ & $0.84{\pm}0.10$ & $0.87{\pm}0.23$ & $\mathbf{1.35{\pm}0.20}$ & $\mathbf{1.57{\pm}0.62}$ \\
\quad +dom & $\mathbf{0.998}$ & $\mathbf{0.41{\pm}0.04}$ & $\mathbf{0.39{\pm}0.01}$ & $\mathbf{0.71{\pm}0.07}$ & $1.55{\pm}0.17$ & $\mathbf{0.996}$ & $\mathbf{0.67{\pm}0.09}$ & $\mathbf{0.80{\pm}0.25}$ & $2.29{\pm}0.79$ & $5.87{\pm}1.79$ \\
\midrule
 & \multicolumn{5}{c}{\textit{SK-Grid (full)}} & \multicolumn{5}{c}{\textit{SK-Grid (L-shape)}} \\
\cmidrule(lr){2-6}\cmidrule(lr){7-11}
Chinchilla & $0.992$ & $0.81{\pm}0.14$ & $0.83{\pm}0.11$ & $1.44{\pm}0.03$ & $5.17{\pm}0.28$ & $0.955$ & $2.19{\pm}0.10$ & $6.09{\pm}0.24$ & $3.63{\pm}0.13$ & $14.63{\pm}0.39$ \\
\quad +dom & $0.991$ & $0.80{\pm}0.19$ & $0.59{\pm}0.05$ & $1.38{\pm}0.13$ & $3.67{\pm}0.20$ & $0.971$ & $1.70{\pm}0.15$ & $4.96{\pm}0.39$ & $3.07{\pm}0.20$ & $13.07{\pm}0.60$ \\
\ska{} & $\mathbf{0.998}$ & $0.33{\pm}0.11$ & $\mathbf{0.39{\pm}0.05}$ & $0.58{\pm}0.07$ & $0.70{\pm}0.39$ & $\mathbf{0.998}$ & $\mathbf{0.33{\pm}0.03}$ & $\mathbf{0.77{\pm}0.44}$ & $\mathbf{0.55{\pm}0.08}$ & $\mathbf{1.16{\pm}0.53}$ \\
\quad +dom & $\mathbf{0.998}$ & $\mathbf{0.32{\pm}0.12}$ & $0.46{\pm}0.05$ & $\mathbf{0.45{\pm}0.05}$ & $\mathbf{0.41{\pm}0.21}$ & $0.996$ & $0.54{\pm}0.19$ & $2.08{\pm}0.94$ & $0.66{\pm}0.27$ & $2.99{\pm}2.05$ \\
\midrule
 & \multicolumn{5}{c}{\textit{Farseer-code (full)}} & \multicolumn{5}{c}{\textit{Farseer-code (L-shape)}} \\
\cmidrule(lr){2-6}\cmidrule(lr){7-11}
Chinchilla & $0.998$ & $0.28{\pm}0.07$ & $0.93{\pm}0.16$ & $0.60{\pm}0.04$ & -- & $0.991$ & $0.59{\pm}0.06$ & $1.93{\pm}0.18$ & $1.60{\pm}0.12$ & -- \\
\quad +dom & $0.998$ & $0.29{\pm}0.06$ & $0.80{\pm}0.06$ & $0.65{\pm}0.05$ & -- & $0.991$ & $0.61{\pm}0.06$ & $1.99{\pm}0.09$ & $1.60{\pm}0.11$ & -- \\
\ska{} & $\mathbf{0.999}$ & $0.24{\pm}0.09$ & $0.67{\pm}0.19$ & $0.26{\pm}0.06$ & -- & $0.993$ & $0.51{\pm}0.11$ & $\mathbf{1.51{\pm}0.59}$ & $\mathbf{1.23{\pm}0.50}$ & -- \\
\quad +dom & $\mathbf{0.999}$ & $\mathbf{0.23{\pm}0.09}$ & $\mathbf{0.60{\pm}0.12}$ & $\mathbf{0.24{\pm}0.06}$ & -- & $\mathbf{0.994}$ & $\mathbf{0.50{\pm}0.15}$ & $1.57{\pm}0.43$ & $1.27{\pm}0.41$ & -- \\
\midrule
 & \multicolumn{5}{c}{\textit{Chinchilla (full)}} & \multicolumn{5}{c}{} \\
\cmidrule(lr){2-6}
Chinchilla & $\mathbf{0.993}$ & $0.63{\pm}0.15$ & $\mathbf{1.16{\pm}0.33}$ & $0.63{\pm}0.06$ & -- &  &  &  &  &  \\
\quad +dom & $0.991$ & $\mathbf{0.59{\pm}0.26}$ & $1.52{\pm}0.14$ & $0.47{\pm}0.02$ & -- &  &  &  &  &  \\
\ska{} & $\mathbf{0.993}$ & $0.61{\pm}0.14$ & $1.28{\pm}0.29$ & $0.51{\pm}0.03$ & -- &  &  &  &  &  \\
\quad +dom & $0.990$ & $0.64{\pm}0.25$ & $1.43{\pm}0.24$ & $\mathbf{0.44{\pm}0.03}$ & -- &  &  &  &  &  \\
\bottomrule
\end{tabular}
\end{table}

\end{document}